\documentclass[runningheads]{llncs}
\usepackage[T1]{fontenc}
\newcommand{\inlineRoot}[1]{\scalebox{0.8}{$\sqrt{#1}$}}
\usepackage{booktabs}
\usepackage{newtxtext}       %
\usepackage[varvw]{newtxmath}       
\usepackage{multirow}
\usepackage[hidelinks]{hyperref}
\usepackage{graphicx}

\usepackage{orcidlink}
\renewcommand{\orcidID}[1]{\orcidlink{#1}}

\makeatletter
\newcommand{\appendixsetup}{%
  \renewcommand{\thetable}{A\arabic{table}}\setcounter{table}{0}%
  \renewcommand{\thefigure}{A\arabic{figure}}\setcounter{figure}{0}}
\makeatother

\usepackage[misc]{ifsym}
\newcommand{\corrauthor}{\scalebox{0.7}{\raisebox{1.3ex}{(\Letter)}}}
\begin{document}
\title{costream: A Streaming Evaluation Protocol for Wearable Fall Detection}
%
%
%
\author{Timilehin B. Aderinola\inst{1}\corrauthor\orcidID{0000-0002-4770-5871} \and
Ilaria D'Ascanio\inst{3}\orcidID{0009-0000-1568-7149} \and
Luca Palmerini\inst{3}\orcidID{0000-0003-4758-662X} \and
Lorenzo Chiari\inst{3}\orcidID{0000-0002-2318-4370} \and
Jochen Klenk\inst{4}\orcidID{0000-0002-5987-447X} \and
Clemens Becker\inst{4,5}\orcidID{0000-0003-1624-8353} \and
Brian Caulfield\inst{2}\orcidID{0000-0003-0290-9587} \and
Georgiana Ifrim\inst{1}\orcidID{0000-0002-8400-2972}}

\authorrunning{T.B. Aderinola et al.}

\institute{School of Computer Science, University College Dublin (UCD), Ireland
\and
UCD School of Public Health, Physiotherapy and Sports Science, Ireland\\
\email{\{timilehin.aderinola,georgiana.ifrim,b.caulfield\}@ucd.ie}
\and
University of Bologna, Italy\\
\email{\{luca.palmerini,ilaria.dascanio2,lorenzo.chiari\}@unibo.it}
\and
Robert Bosch Hospital, Stuttgart, Germany\\
\email{\{jochen.klenk,clemens.becker\}@rbk.de}
\and
Heidelberg University Hospital, Heidelberg, Germany}
\maketitle              
%
\begin{abstract}
Wearable fall detectors are commonly evaluated on presegmented windows, where a fall window is extracted at an annotated impact and non-fall windows are sampled from fall-free signal. However, a deployed detector processes a continuous sensor stream in which the location of a fall is unknown. We specify a streaming evaluation protocol in which a detector traverses an unsegmented recording, aggregates window-level scores into a temporal confidence map, and emits timestamped events that are evaluated against the annotated fall. Holding the dataset, participant-wise splits, window construction and fitted models fixed, we compare presegmented and streaming evaluation on clinically verified real-world falls. The two protocols identify the same best-performing method overall, but select different methods in 6 of 20 model-window configurations, with the divergence associated with differences in the temporal clustering of false alarms. Streaming evaluation also exposes deployment-relevant quantities, including false alarm rate over elapsed monitoring time, detection delay, and the effect of the decision threshold, that are not captured by per-window evaluation. These results show that evaluation protocol can materially affect conclusions about wearable fall detectors and should be considered when assessing their suitability for continuous monitoring.
\end{abstract}

\section{Introduction}
\label{sec:introduction}

Falls are the second leading cause of unintentional injury death worldwide, accounting for an estimated 684,000 deaths annually, with adults over 60 at highest risk~\cite{step_safely_2021,vaishya2020falls}. The consequences of a fall also depend on what happens after the impact, as remaining on the ground without assistance can increase the risk of adverse outcomes~\cite{schwickert2017reading}. Automatic fall detection can therefore support timely intervention by reducing the time between a fall and the initiation of a response.

Wearable inertial sensors are well suited to this task because they are inexpensive, unobtrusive, and can continuously monitor movement during daily activities~\cite{mohan2024artificial}. A detector deployed in this setting receives a continuous accelerometer stream and must determine when a fall has occurred. However, evaluation practice often does not reflect this setting. The prevailing protocol extracts a window anchored to an annotated impact, draws additional windows from fall-free signal, and evaluates how accurately a classifier separates the two~\cite{palmerini2020accelerometer,aderinola2024accurate,resnet}. This evaluates recognition of a fall given its approximate location rather than detection in a stream where the location of a fall is unknown. The practice is partly inherited from the structure of commonly used datasets, since many studies use simulated falls recorded as discrete trials~\cite{sucerquia2017sisfall,saleh2020fallalld}, where each trial is already a segment.

We therefore specify an evaluation protocol in which a detector processes a continuous recording without prior knowledge of the fall location and converts window-level predictions into timestamped events. The resulting detections are evaluated using an event-level any-overlap criterion, following approaches used in other continuous-monitoring domains~\cite{ward2011performance,mesaros2016metrics,dan2024szcore}. To isolate the effect of evaluation protocol, we hold the dataset, participant-wise splits, window construction, and fitted models fixed. We then compare the conventional presegmented evaluation with the streaming protocol and examine how their conclusions differ. Our contributions are as follows.

\begin{enumerate}
\item We formulate wearable fall detection as continuous event detection, aggregating overlapping window predictions into a temporal confidence map from which timestamped detection events are extracted. The protocol is independent of the underlying window classifier and enables deployment-relevant analysis of false alarm rate, detection delay and threshold-dependent operating points.

\item By empirically comparing conventional presegmented and streaming evaluation using the same real-world recordings, participant-wise splits, window sizes and fitted models, we show that the evaluation protocols can lead to different model selections and relate these differences to the temporal clustering of false alarms.

\item We release \texttt{costream}, an open-source implementation of the protocol, along with the analyses reported in this study.\footnote{\url{https://github.com/mlgig/costream}}
\end{enumerate}

The remainder of the paper is organised as follows. Section~\ref{sec:related-work} reviews related work on wearable fall detection and event-level evaluation. Section~\ref{sec:protocol} describes the streaming evaluation protocol and its comparison with presegmented evaluation. Section~\ref{sec:experiments} presents the experimental setup and results, including the cross-protocol comparison and operating-point analysis. Section~\ref{sec:discussion} discusses the findings and limitations, and Section~\ref{sec:conclusion} concludes the paper.
 
\section{Related Work}
\label{sec:related-work}
\subsection{Wearable Fall Detection and Evaluation Practice}
\label{sec:rw-detection}

Wearable fall detection methods include threshold-based, machine learning-based, and hybrid approaches~\cite{rastogi2021systematic}. Most, however, evaluate a window of sensor data presented in isolation and labelled as either a fall or an activity of daily living. This formulation is convenient for comparing classifiers, but does not directly evaluate whether a detector can locate a fall in a continuous recording.

Real-world falls are rare and difficult to record~\cite{klenk2016farseeing}, so much of the literature relies on simulated datasets such as SisFall~\cite{sucerquia2017sisfall} and FallAllD~\cite{saleh2020fallalld}, which are collected as discrete trials and naturally support per-window classification. Even studies using real-world falls from FARSEEING commonly extract windows around the annotated impact and report per-window performance~\cite{palmerini2020accelerometer,aderinola2024accurate,resnet}. Such evaluation measures recognition of a fall at a known location, but does not quantify how frequently a detector would raise alarms during the surrounding periods of ordinary activity.

\subsection{Event-Level Evaluation in Continuous Monitoring}
\label{sec:rw-event-detection}

Event-level evaluation has been developed in other continuous-monitoring domains where events are sparse and timing matters. Activity recognition and sound event detection distinguish event-level from segment-based scoring and account for temporal offsets~\cite{ward2011performance,mesaros2016metrics}. Similar issues arise in EEG seizure detection, where event-based scoring can lead to different conclusions from sample-based evaluation~\cite{dan2024szcore}. Recent benchmarking work has therefore advocated standardised event-level evaluation of continuous detectors~\cite{samanos2025benchmark}.

We adopt this event-level perspective for wearable fall detection and evaluate detections against the annotated fall using an any-overlap criterion with an asymmetric tolerance. We then ask whether conclusions obtained from conventional presegmented evaluation are preserved when the same models are evaluated as continuous detectors.
 
\section{Streaming Evaluation Protocol}
\label{sec:protocol}
 
\subsection{Dataset}
\label{sec:dataset}

We evaluate on the FARSEEING real-world fall repository~\cite{klenk2016farseeing}, a collection of clinically verified falls recorded during everyday activity. Each recording contains twenty minutes of continuous signal with an annotated impact near its midpoint. We use the subset recorded with sensors at the fifth lumbar vertebra, comprising 145 falls from 41 participants. Two recordings were captured at 20~Hz and resampled to 100~Hz; the remainder were recorded at 100~Hz.

\subsection{Preprocessing and Segmentation}
\label{sec:preprocessing}

\begin{figure}[t]
\centering
\includegraphics[width=0.8\linewidth]{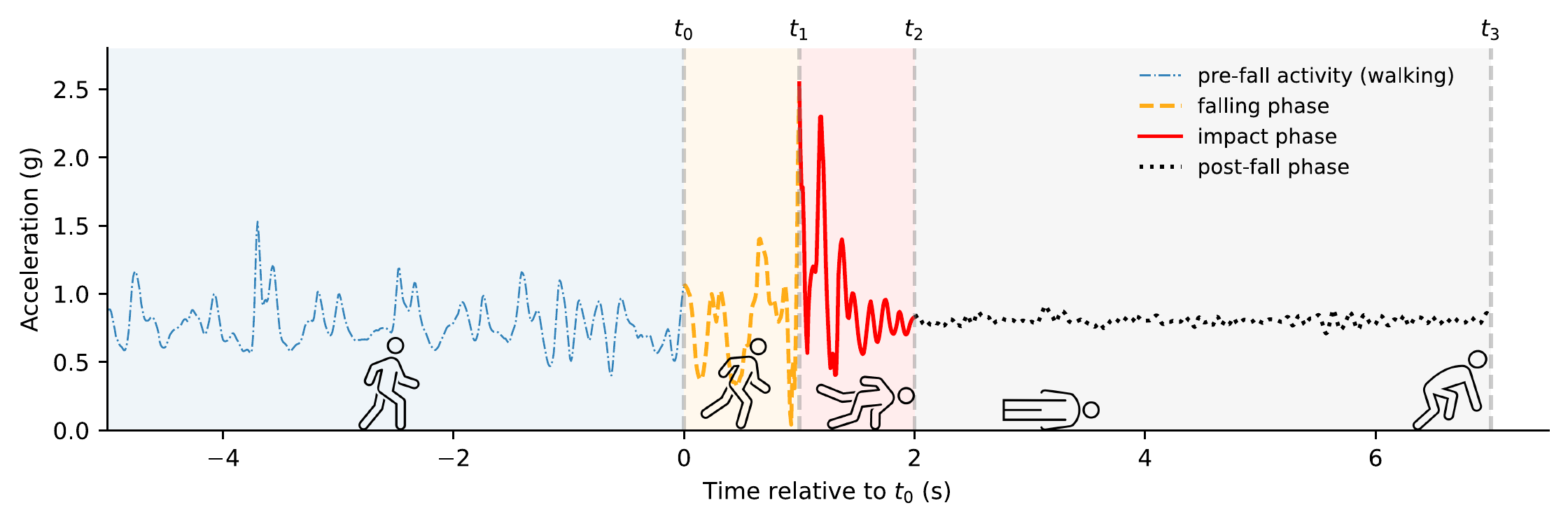}
\caption{A multiphase fall from FARSEEING, showing the falling, impact, and post-fall phases. The post-fall phase covers the resting period during which the faller remains on the ground, followed by recovery.}
\label{fig:multiphase}
\end{figure}

The tri-axial signals are reduced to acceleration magnitude $M = \inlineRoot{\,\mathit{Acc}_x^2 + \mathit{Acc}_y^2 + \mathit{Acc}_z^2\,}$, which is robust to sensor orientation. Windows are standardised before classification. Following Becker et al.~\cite{becker2012proposal} and Palmerini et al.~\cite{palmerini2020accelerometer}, each fall window of length $w$ comprises a one-second falling phase, a one-second impact phase, and a post-fall phase of $w-2$ seconds (Fig.~\ref{fig:multiphase}). The annotated impact index $f$ marks the start of the impact phase, so the positive window is $[f-1,f+w-1)$.

Windows are generated with a one-second step. One positive window is extracted per fall, while negative windows are sampled from portions of the recording outside
$[f-(w+20),f+20)$ to avoid ambiguous signal surrounding the fall. The margin is used for labelling only and is unrelated to the scoring interval in Section~\ref{sec:scoring}. A window is admissible if its maximum acceleration magnitude between one and two seconds from its start exceeds a configurable threshold $\theta_g$. Windows failing this condition are excluded from the negative training pool and assigned a score of zero at inference. The value of $\theta_g$ is specified as part of the experimental configuration. No features are extracted; each $w$-second window is presented as a raw vector of length $T=100w$.
 
\subsection{Event Detection}
\label{sec:formulation}

\begin{figure}[t]
    \centering
    \includegraphics[width=0.8\linewidth]{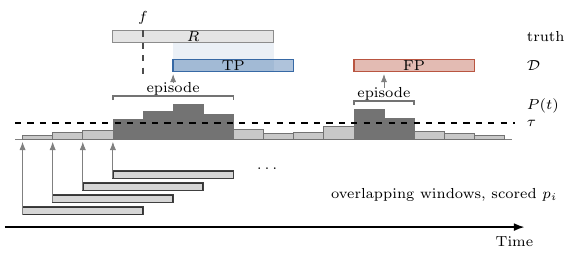}
    \caption{Overlapping windows are scored independently and aggregated by pointwise maximum into a temporal confidence map $P(t)$. Thresholding at $\tau$ yields detection episodes, each emitting one detection. A detection overlapping the reference interval $R$ counts as a true positive; one with no overlap is a false positive.}
    \label{fig:pipeline}
\end{figure}

The detector maps a continuous accelerometer recording to timestamped detection events through four stages: window scoring, temporal aggregation, thresholding, and event extraction (Fig.~\ref{fig:pipeline}). Only window scoring depends on the classifier; the remaining stages are fixed by the protocol.

Let $\mathbf{x}=(x_1,\dots,x_L)$ denote the acceleration magnitude signal sampled at $f_s=100$~Hz. The signal is traversed using $w$-second windows with a one-second step: for window $i$, $X_i=(x_{s_i+1},\dots,x_{s_i+n_w})$ with $s_i=(i-1)f_s$ and $n_w=wf_s$. A window is admissible if its peak magnitude during the impact phase exceeds the threshold $\theta_g$,
\[
\gamma(X_i)=\mathbf{1}\!\left[\max_{u\,\in\,[f_s,\,2f_s)}X_i(u)>\theta_g\right],
\]
where $u$ indexes samples within the window, so the maximum is taken over the interval one to two seconds from its start. Given a trained classifier $h:\mathbb{R}^{n_w}\rightarrow[0,1]$ applied to the standardised window, the window score is $p_i=\gamma(X_i)h(X_i)$. The annotated fall location is not used during prediction. Because adjacent windows overlap, multiple scores correspond to each instant. We aggregate them using a pointwise maximum:

\begin{equation}
\label{eq:confidence-map}
P(t)=\max_{i\in\mathcal{C}(t)}p_i,
\qquad
P(t)=0\ \text{if }\mathcal{C}(t)=\emptyset,
\end{equation}

\noindent where $\mathcal{C}(t)$ is the set of windows containing $t$. This introduces no learned parameters and preserves the signal time base. Thresholding gives $D(t)=\mathbf{1}[P(t)>\tau]$. Maximal contiguous positive regions form detection episodes, and a debounce step emits one detection per episode, merging episodes separated by less than 60 seconds so that a sustained burst of firings raises a single alarm. The merge interval is set long relative to the scoring tolerance because a monitoring service receiving several alerts within a minute experiences one event. The emitted interval is the window with the earliest start among those attaining the maximum score within the episode: $d_k=[s_{i_k},s_{i_k}+n_w)$. The resulting output is a set $\mathcal{D}=\{d_1,\dots,d_K\}$ of timestamped intervals determined by the recording and detector rather than by a predefined segmentation.

\subsection{Event Scoring}
\label{sec:scoring}

Each recording provides an annotated impact index $f$. Because the clinically relevant fall event extends beyond the one-second impact annotation, we define the reference interval as $R=[f-1,f+t)$, where $t=20~\text{s}$. The one-second pre-impact interval corresponds to the falling phase used in training, while the 20-second post-impact interval covers the recovery period. Inability to recover within 24.5~s has been associated with more serious outcomes~\cite{schwickert2017reading}.

A detected interval $d\in\mathcal{D}$ is a true positive if $d\cap R\neq\emptyset$ and a false positive otherwise. A fall with no overlapping detection is a false negative. Thus, we use an event-level any-overlap criterion. Since one detection is emitted per episode, multiple overlapping windows within the same episode cannot produce duplicate true positives.

Precision, recall and $F_1$ are computed from pooled event counts within each fold. For matched detections, detection delay is the signed difference between the detection window start and the annotated impact, with negative values indicating detection before impact. We report its distribution rather than only its mean.

\subsection{Two Evaluation Protocols}
\label{sec:two-protocols}

We compare two ways of evaluating the same fitted window classifiers on the same recordings. Both use the annotated impact index $f$ for evaluation, but it enters at different stages.

\paragraph{Protocol A: presegmented evaluation.}
A positive window is extracted at $[f-1,f+w-1)$ and negative windows are sampled from the remaining recording, excluding the interval around the fall. The classifier is applied independently to these windows and performance is measured using per-window precision, recall and $F_1$. Thus, $f$ determines the evaluation windows before prediction.

\paragraph{Protocol B: streaming evaluation.}
The recording is traversed from an arbitrary starting point using the mapping in Section~\ref{sec:formulation}. The resulting detections are scored against the reference interval $R$ using event-level precision, recall and $F_1$. Here, $f$ is used only after prediction, to score the emitted detections.

The protocols therefore measure different quantities. Protocol~A evaluates recognition of a fall at a known location, whereas Protocol~B evaluates whether a detector can locate falls in a continuous recording. Because their units of evaluation differ, we do not compare their absolute $F_1$ values. Instead, we compare whether they select the same models across the same recordings, folds and window sizes.

\subsection{Selecting an Operating Point}
\label{sec:operating-point}

In streaming evaluation, the threshold $\tau$ is applied to the temporal confidence map and therefore determines the number of alarm episodes as well as fall detection performance. We treat threshold selection as an operating-point choice rather than as a contribution of the classifier. Cost-sensitive threshold selection is standard~\cite{elkan2001foundations}.

Let $c=C_{\mathrm{FN}}/C_{\mathrm{FP}}$ denote the relative cost of a missed fall. We define
\begin{equation}
\label{eq:gain}
g(\tau)=-
\left(\mathrm{FP}(\tau)+c\,\mathrm{FN}(\tau)\right),
\end{equation}
\noindent where FP and FN are event-level counts. We evaluate 100 evenly spaced thresholds over $[0,1]$ using five-fold cross-validation on the development set and select the threshold with the highest mean gain. The cost ratio is treated as a deployment parameter. We therefore examine its effect in Section~\ref{sec:sweeps}; $c=2$ is used where a single value is required for illustration.

\section{Experiments}
\label{sec:experiments}

\subsection{Setup}
\label{sec:setup}

All data splits are participant-wise. Nine of the 41 participants, contributing 21 recordings, are held out as an untouched test cohort. Results on this cohort are reported in Appendix~\ref{app:holdout}. The remaining 32 participants and 124 falls form the development set used for five-fold participant-wise cross-validation. Within each fold and window size, the same fitted model is evaluated under both protocols at $\tau=0.5$. Threshold selection is examined separately in Section~\ref{sec:sweeps}.

We evaluate five classifiers spanning distinct families (Table~\ref{tab:model-overview}), using the default parameters of their respective implementations. No feature extraction is applied. Each model is trained at four window sizes, $w \in \{3, 10, 30, 60\}$ s, in each of five folds, giving 20 model-window configurations per classifier. The same configurations are evaluated under both protocols.

\begin{table}[t]
\centering
\setlength{\tabcolsep}{6pt}
\caption{Classifiers evaluated.}
\label{tab:model-overview}
\begin{tabular}{lll}
\toprule
Model & Family & Implementation \\
\midrule
Catch22~\cite{lubba2019catch22}      & Feature-based & \texttt{aeon} \\
ExtraTrees~\cite{pedregosa2011scikit} & Tree-based    & \texttt{scikit-learn} \\
MiniRocket~\cite{dempster2021minirocket} & Convolutional & \texttt{aeon} \\
QUANT~\cite{dempster2024quant}       & Interval-based & \texttt{aeon} \\
ResNet~\cite{wang2017time}           & Deep CNN      & \texttt{aeon} \\
\bottomrule
\end{tabular}
\end{table}

The admissibility threshold is fixed at 1.4~g, following Palmerini et al.~\cite{palmerini2020accelerometer}, who report acceleration peaks above this level in recorded real-world falls. The threshold is configurable in \texttt{costream} but is held fixed here so that it does not vary between models or evaluation protocols. False alarm rates are averaged across the four window-size configurations within each model.
 
Experiments were run with Python 3.10.18 on a Linux server (Ubuntu 22.04.3 LTS) with 1.5~TB RAM and an NVIDIA GeForce RTX 4090 GPU (24~GB), using \texttt{aeon} v1.2.0 and \texttt{scikit-learn} v1.6.1.

\subsection{Cross-Protocol Comparison}
\label{sec:cross-protocol}
\label{sec:mechanism}

Table~\ref{tab:aggregate} reports mean $F_1$ for each model under the two protocols. QUANT ranks first under both, showing that presegmented evaluation does identify the best method overall on this dataset. However, agreement below the top-ranked model is weaker. ExtraTrees and MiniRocket are nearly tied under Protocol~A ($0.683$ vs.\ $0.682$) but are separated by $0.071$ under Protocol~B, while ResNet moves from fifth to fourth.

Across the 20 model-window configurations, the protocols select different best models in 6 cases. In 5 of these, Protocol~A selects QUANT whereas Protocol~B selects MiniRocket or ResNet. Pairwise inversions are most common for ExtraTrees--ResNet (9/20) and MiniRocket--QUANT (6/20), while QUANT does not invert with ExtraTrees or Catch22. Wilcoxon tests across the 20 model-window configurations give $p=0.002$ for ResNet, $0.021$ for ExtraTrees and $0.048$ for MiniRocket, compared with $p=0.152$ for QUANT and $0.075$ for Catch22.

\begin{table}[t]
\centering
\setlength{\tabcolsep}{6pt}
\caption{Mean $F_1$ per model under each protocol over the 20 model-window configurations, with the implied rank. The two columns are not commensurable; they are placed side by side to compare orderings, not values.}
\label{tab:aggregate}
\begin{tabular}{lcccccc}
\toprule
& \multicolumn{3}{c}{Protocol A (per-window)} & \multicolumn{3}{c}{Protocol B (event-level)} \\
\cmidrule(lr){2-4} \cmidrule(lr){5-7}
Model & Mean $F_1$ & SD & Rank & Mean $F_1$ & SD & Rank \\
\midrule
QUANT      & 0.822 & 0.087 & 1 & 0.811 & 0.074 & 1 \\
ExtraTrees & 0.683 & 0.097 & 2 & 0.668 & 0.095 & 3 \\
MiniRocket & 0.682 & 0.186 & 3 & 0.739 & 0.144 & 2 \\
Catch22    & 0.541 & 0.225 & 4 & 0.563 & 0.211 & 5 \\
ResNet     & 0.527 & 0.216 & 5 & 0.638 & 0.160 & 4 \\
\bottomrule
\end{tabular}
\end{table}

The direction of the protocol difference is related to the temporal clustering of false alarms. Protocol A counts each misclassified window separately, whereas Protocol B emits one false alarm per detection episode. We therefore define the \textit{collapse ratio} as the number of window-level false firings per emitted false alarm episode. Table 3 reports it alongside the protocol gap, which measures how much a model's score changes between protocols, and the Wilcoxon $p$-value, which measures how consistently it changes across the 20 configurations. The gap broadly increases with the collapse ratio. The five methods separate into two groups. QUANT, Catch22, and ExtraTrees have ratios close to one and are scored similarly by both protocols, whereas MiniRocket and ResNet fire repeatedly within single regions of activity and show the two largest positive gaps. The intervals in Table 3 do not resolve the ordering within either group.

The collapse ratio is not solely a property of the method. Some methods, ResNet in particular, tend to fire repeatedly within a single region of fall-like activity, but the degree to which they do so is also influenced substantially by the recordings being evaluated and by the window size. Episode-level counting, implemented through a debounce step, is available only to Protocol~B. Episodes are contiguous runs of a confidence map defined over a continuous recording, and a presegmented window set has neither contiguity nor episodes. The streaming results are also sensitive to the scoring interval. Expanding the pre-impact acceptance boundary from $f-1$ to $f-(w+20)$ changes $F_1$ in 12 of 100 cells, always upward and by up to $0.065$. MiniRocket and ResNet account for 8 of these 12 changes, whereas ExtraTrees is unaffected. Thus, scoring choices can alter results by an amount comparable to differences between adjacent models, particularly for models with more frequent firings.

\begin{table}[t]
\centering
\caption{False alarm clustering and the protocol gap. The collapse ratio is window-level false firings per emitted false alarm episode, with a 95\% participant-cluster bootstrap interval over 4000 resamples. The ratio has a floor of 1.0 by construction. The gap is mean $F_1$ under Protocol~B minus Protocol~A, reported to show how it varies across models rather than as a difference in performance.}
\label{tab:clustering}
\begin{tabular}{lcccc}
\toprule
Model & Collapse ratio & 95\% CI & Gap (B $-$ A) & Wilcoxon $p$ \\
\midrule
QUANT      & 1.09 & [1.00, 1.24] & $-0.011$ & 0.152 \\
Catch22    & 1.33 & [1.05, 1.46] & $+0.022$ & 0.075 \\
ExtraTrees & 1.38 & [1.00, 1.69] & $-0.015$ & 0.021 \\
MiniRocket & 2.86 & [2.05, 3.85] & $+0.057$ & 0.048 \\
ResNet     & 6.07 & [2.50, 9.94] & $+0.111$ & 0.002 \\
\bottomrule
\end{tabular}
\end{table}
 
\subsection{Operating Points}
\label{sec:sweeps}

\begin{figure}[t]
\centering
\includegraphics[width=\linewidth]{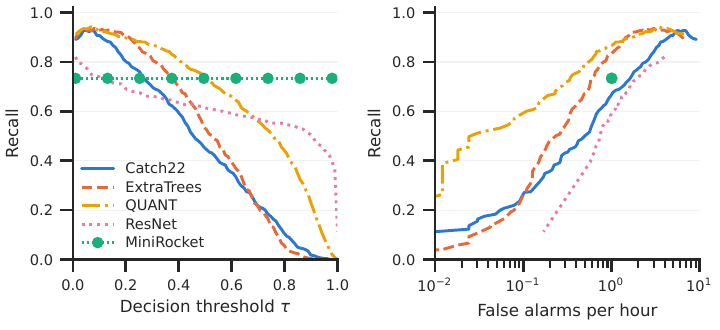}
\caption{Recall as a function of decision threshold (left) and resulting false alarm rate (right), averaged across the four window-size configurations within each model. The right-hand panel has no equivalent under presegmented evaluation because the sampled window pool has no elapsed-time denominator. MiniRocket is shown as a single operating point because its default estimator provides hard decisions rather than probabilities.}
\label{fig:operating-curves}
\end{figure}

Because $\tau$ is applied to the temporal confidence map, each fitted model defines a family of operating points with different recall and false alarm rates. All quantities in this section are derived from the cached window-level scores, with no refitting. Figure~\ref{fig:operating-curves} shows recall against both threshold and false alarm rate. The latter is a deployment-relevant measure because it uses elapsed monitoring time, which is unavailable from a presegmented window pool. At $\tau = 0.5$, QUANT achieves a recall of 0.748 at a mean rate of 0.35 false alarms per hour; at $\tau = 0.07$, recall rises to 0.94 at 3.41 per hour.

QUANT achieves the highest recall across most of the observed false alarm range, so the ranking at $\tau=0.5$ is not overturned. The curves also show that the separation between methods depends on the allowed alarm burden. The methods are closer above roughly three false alarms per hour and more widely separated below one per hour. A fixed-threshold comparison therefore captures only one operating point and does not show how performance changes as the alarm burden is varied.

MiniRocket appears as a single operating point because its default estimator returns hard decisions rather than calibrated probabilities. Thresholding therefore does not change its predictions. We retain the library default for comparability and do not calibrate MiniRocket separately, since doing so would introduce model-specific tuning into an otherwise fixed comparison. Its operating point is recall $0.734$ at $1.00$ false alarms per hour.

Table~\ref{tab:cost-sweep} reports the threshold selected by maximising the gain function of Section~\ref{sec:operating-point} for $c\in\{1,2,5,10\}$, together with the resulting operating point. The $c=1$ row shows the effect of threshold selection without cost asymmetry. MiniRocket is excluded because its hard predictions provide no threshold-dependent operating points.
 
\begin{table}[t]
\centering
\setlength{\tabcolsep}{6pt}
\caption{Threshold selected by cost-sensitive optimisation and the resulting operating point. $c$ is the assumed cost of a missed fall relative to a false alarm; $\tau$ is the mean threshold selected across folds. FA/h is false alarm episodes per hour of recording. MiniRocket is excluded: its scores are binary, so the gain function is constant in $\tau$ and cost-based selection cannot act.}
\label{tab:cost-sweep}
\begin{tabular}{llccccc}
\toprule
Model & $c$ & $\tau$ & Precision & Recall & $F_1$ & FA/h \\
\midrule
\multirow{5}{*}{Catch22}
 & default & 0.50 & 0.782 & 0.456 & 0.576 & 0.38 \\
 & 1  & 0.40 & 0.718 & 0.595 & 0.650 & 0.70 \\
 & 2  & 0.33 & 0.657 & 0.675 & 0.666 & 1.06 \\
 & 5  & 0.20 & 0.487 & 0.782 & 0.600 & 2.47 \\
 & 10 & 0.14 & 0.423 & 0.881 & 0.572 & 3.60 \\
\midrule
\multirow{5}{*}{ExtraTrees}
 & default & 0.50 & 0.847 & 0.536 & 0.657 & 0.29 \\
 & 1  & 0.34 & 0.739 & 0.776 & 0.757 & 0.82 \\
 & 2  & 0.28 & 0.680 & 0.831 & 0.748 & 1.17 \\
 & 5  & 0.18 & 0.627 & 0.919 & 0.746 & 1.64 \\
 & 10 & 0.18 & 0.627 & 0.919 & 0.746 & 1.64 \\
\midrule
\multirow{5}{*}{QUANT}
 & default & 0.50 & 0.865 & 0.748 & 0.802 & 0.35 \\
 & 1  & 0.44 & 0.829 & 0.784 & 0.806 & 0.48 \\
 & 2  & 0.36 & 0.789 & 0.843 & 0.815 & 0.68 \\
 & 5  & 0.28 & 0.706 & 0.875 & 0.781 & 1.09 \\
 & 10 & 0.19 & 0.614 & 0.891 & 0.727 & 1.68 \\
\midrule
\multirow{5}{*}{ResNet}
 & default & 0.50 & 0.623 & 0.613 & 0.618 & 1.11 \\
 & 1  & 0.82 & 0.663 & 0.540 & 0.596 & 0.82 \\
 & 2  & 0.27 & 0.570 & 0.688 & 0.623 & 1.55 \\
 & 5  & 0.08 & 0.468 & 0.756 & 0.578 & 2.58 \\
 & 10 & 0.03 & 0.412 & 0.794 & 0.542 & 3.41 \\
\bottomrule
\end{tabular}
\end{table}

The cost sweep illustrates two effects. First, the same recall target can require substantially different alarm burdens. At $c=10$, Catch22 reaches $0.881$ recall at $3.60$ false alarms per hour, whereas QUANT reaches $0.891$ at $1.68$, roughly half the alarm burden for the same detection rate. Second, increasing the cost ratio does not necessarily continue to change the operating point. ExtraTrees selects the same threshold at $c=5$ and $c=10$, with recall remaining at $0.919$. The $c=1$ results also show that part of the improvement over the default threshold comes from threshold selection itself rather than from imposing an asymmetric cost.
 
\section{Discussion}
\label{sec:discussion}

The two evaluation protocols identified the same best-performing method on the FARSEEING dataset but disagreed in 6 of 20 model-window configurations. The difference is related to how false alarms are counted: Protocol~A charges each misclassified window separately, whereas Protocol~B charges one false positive per emitted alarm episode. Models with more temporally clustered false alarms therefore show larger differences between protocols.

The results also show why streaming evaluation provides information that is not captured by presegmented scoring. False alarm rate requires elapsed monitoring time, detection delay requires detections placed on a timeline, and threshold sweeps expose the trade-off between detection and alarm burden. These quantities become directly measurable once window scores are converted into a temporal confidence map over the recording, and they are the quantities a constrained deployment is budgeted against. On a wearable device, the alarm rate determines how often the system must transmit or wake a downstream service. Since the protocol is independent of the classifier, a compressed or quantised model intended for on-device inference is evaluated in exactly the same way, though we do not evaluate such models here. The agreement on the top-ranked model also indicates that protocol choice does not necessarily change the overall winner, but can change conclusions among competing methods.

Several limitations bound these findings. The evaluation uses one dataset in which each 20-minute recording contains a single fall near its midpoint, so reported false alarm rates may not represent free-living monitoring. The clustering analysis covers five methods and should be regarded as illustrative rather than a general model property. We also evaluate each method at its library defaults, so differences reflect the combined effect of classifier and implementation rather than representations in isolation. Finally, the event-scoring criterion contains design choices. Our sensitivity analysis shows that changing the pre-impact boundary can alter results.
 
\section{Conclusion}
\label{sec:conclusion}

We formulated wearable fall detection as a continuous event-detection task and compared presegmented and streaming evaluation while holding the dataset, participant-wise splits, window construction and fitted models fixed. The protocols identified the same best-performing method on this dataset but selected different methods in 6 of 20 model-window configurations, with larger differences associated with more temporally clustered false alarms. Streaming evaluation also exposes deployment-relevant quantities, including false alarm rate over elapsed monitoring time, detection delay and threshold-dependent operating points, that are not captured in the same way by presegmented evaluation. We release \texttt{costream} to support application of the protocol to other continuous event detection datasets.
 
\begin{credits}
\subsubsection{\ackname}
This study has emanated from research funded by Research Ireland under the Government of Ireland Postdoctoral Fellowship Programme (Grant Number: GOIPD/2025/1434).
 
\subsubsection{\discintname}
The authors have no competing interests to declare.
\end{credits}
 
\bibliographystyle{splncs04}
\bibliography{refs}


\clearpage
\section*{Appendix}
\appendix
\appendixsetup
\section{Held-Out Cohort Evaluation}
\label{app:holdout}

Section 4 reports results from five-fold cross-validation on the 32 development participants. This appendix evaluates the same protocols on the nine participants and 21 recordings held out from all earlier analysis.

\subsection{Setup}

Each model is evaluated at a single window size $w^{\ast}$, selected as the window size with the highest mean Protocol B $F_1$ across the five development folds. The selected values are $w^{\ast} = 3$~s for ExtraTrees and ResNet and $w^{\ast} = 10$~s for Catch22, MiniRocket and QUANT. Window sizes were selected on the development folds, so any development figure reported at $w^{\ast}$ below is in-sample and is included only to make the hold-out numbers interpretable. The held-out cohort takes no part in model selection, window size selection or threshold selection.

\subsection{Results}

Table~\ref{tab:holdout} reports both protocols at $\tau = 0.5$ on the held-out cohort. MiniRocket ranks first under both protocols. Two of the five ranks differ between protocols, so the disagreement observed in Section 4.2 is present here in weaker form. The protocols produce identical counts for Catch22 and MiniRocket, which is possible because a detection emitted under Protocol B may correspond to the same window Protocol A classifies.

\begin{table}
\centering
\caption{Held-out cohort, 9 participants and 21 recordings, each model at its own $w^{\ast}$ and $\tau = 0.5$. Counts are given because the cohort contains 21 falls, so a single detection changes recall by approximately $0.05$.}
\label{tab:holdout}
\setlength{\tabcolsep}{5pt}
\begin{tabular}{llrrrrrrr}
\toprule
Model & $w^{\ast}$ & Protocol & TP & FP & FN & Precision & Recall & $F_1$ \\
\midrule
\multirow{2}{*}{MiniRocket} & \multirow{2}{*}{10}
 & A & 17 & 2 & 4 & 0.895 & 0.810 & 0.850 \\
 & & B & 17 & 2 & 4 & 0.895 & 0.810 & 0.850 \\
\midrule
\multirow{2}{*}{ExtraTrees} & \multirow{2}{*}{3}
 & A & 15 & 0 & 6 & 1.000 & 0.714 & 0.833 \\
 & & B & 14 & 0 & 7 & 1.000 & 0.667 & 0.800 \\
\midrule
\multirow{2}{*}{QUANT} & \multirow{2}{*}{10}
 & A & 15 & 2 & 6 & 0.882 & 0.714 & 0.789 \\
 & & B & 16 & 2 & 5 & 0.889 & 0.762 & 0.821 \\
\midrule
\multirow{2}{*}{Catch22} & \multirow{2}{*}{10}
 & A & 14 & 3 & 7 & 0.824 & 0.667 & 0.737 \\
 & & B & 14 & 3 & 7 & 0.824 & 0.667 & 0.737 \\
\midrule
\multirow{2}{*}{ResNet} & \multirow{2}{*}{3}
 & A & 14 & 11 & 7 & 0.560 & 0.667 & 0.609 \\
 & & B & 12 & 9 & 9 & 0.571 & 0.571 & 0.571 \\
\bottomrule
\end{tabular}
\end{table}

\subsection{Transfer of the Cost-Selected Threshold}

The threshold selected on the development folds at $c = 2$ transfers to the held-out cohort without modification (Table~\ref{tab:holdout-cost}). Every model gains recall relative to $\tau = 0.5$, at a cost in precision, which is the trade-off the cost ratio is intended to produce. QUANT reaches a recall of $0.905$ at a precision of $0.826$ on participants that played no part in selecting its threshold. For three of the four models the transferred threshold also improves $F_1$ relative to $\tau = 0.5$. ExtraTrees is the exception, falling from $0.800$ to $0.780$.

\begin{table}[t]
\centering
\caption{Protocol B on the held-out cohort at the development-selected threshold for $c = 2$, applied unchanged. MiniRocket is excluded: its scores are binary, so the gain function is constant in $\tau$.}
\label{tab:holdout-cost}
\setlength{\tabcolsep}{5pt}
\begin{tabular}{llrrrrrrr}
\toprule
Model & $w^{\ast}$ & $\tau$ & TP & FP & FN & Precision & Recall & $F_1$ \\
\midrule
QUANT      & 10 & 0.36 & 19 &  4 &  2 & 0.826 & 0.905 & 0.864 \\
ExtraTrees &  3 & 0.29 & 16 &  4 &  5 & 0.800 & 0.762 & 0.780 \\
Catch22    & 10 & 0.33 & 17 &  7 &  4 & 0.708 & 0.810 & 0.756 \\
ResNet     &  3 & 0.32 & 13 & 11 &  8 & 0.542 & 0.619 & 0.578 \\
\bottomrule
\end{tabular}
\end{table}

\subsection{Clustering and Cohort Composition}

The protocol gaps on the held-out cohort are small, between $0.000$ and $0.037$ in absolute value, and four of the five change sign relative to the development gaps. Two of those apparent reversals are gaps of exactly zero, so the direction is absent rather than reversed.

This is consistent with the collapse ratios at the selected window sizes. At $w^{\ast}$, development collapse ratios range from $1.00$ to $1.78$, against $1.09$ to $6.07$ when pooled across all four window sizes: the clustering reported in Section 4.2 is concentrated at $w = 30$ and $w = 60$, which the selection procedure rejects for every model. At the window sizes development actually selects, there is little clustering in either cohort, and correspondingly little protocol gap.

What the cohort can show is bounded by its size. With 21 falls, one detection moves recall by approximately $0.05$, which exceeds most of the differences in Table~\ref{tab:holdout}; false alarm counts run between zero and eleven per model, low enough that the bootstrap intervals used in Section 4.2 carry no resolution here.

The comparison nonetheless illustrates a dependence the development results alone would not reveal. Clustering behaviour is not a fixed property of a method: it varies with the window size and with which participants are in the cohort. Methods that fire repeatedly within a region of activity are most affected, so their behaviour under either protocol is less stable across cohorts than that of methods whose errors are isolated.
\end{document}